\documentclass{article}

    \usepackage[preprint, nonatbib]{neurips_2024}

\usepackage[utf8]{inputenc} 
\usepackage[T1]{fontenc}    
\usepackage{hyperref}       
\usepackage{url}            
\usepackage{booktabs}       
\usepackage{amsfonts}       
\usepackage{nicefrac}       
\usepackage{microtype}      
\usepackage{xcolor}         
\usepackage{graphicx}
\usepackage[most]{tcolorbox}
\usepackage{enumitem}
\usepackage{colortbl}
\usepackage{multirow}
\usepackage{amssymb}
\usepackage{listings} 

\title{Surgical-LLaVA: Toward Surgical Scenario Understanding via Large Language and Vision Models}
\author{%
    \textbf{Juseong Jin} \\
    Seoul National University Hospital \\
    \texttt{jinju4948@snu.ac.kr} 
    \and
    \textbf{Chang Wook Jeong} \\
    Seoul National University Hospital \\
    \texttt{drboss@snu.ac.kr}
}

\begin{document}

\maketitle

\begin{abstract}
Conversation agents powered by large language models are revolutionizing the way we interact with visual data. Recently, large vision-language models (LVLMs) have been extensively studied for both images and videos. However, these studies typically focus on common scenarios. In this work, we introduce an LVLM specifically designed for surgical scenarios. We integrate visual representations of surgical images and videos into the language feature space. Consequently, we establish a LVLM model, Surgical-LLaVA, fine-tuned on instruction following data of surgical scenarios. Our experiments demonstrate that Surgical-LLaVA exhibits impressive multi-modal chat abilities in surgical contexts, occasionally displaying multi-modal behaviors on unseen instructions. We conduct a quantitative evaluation of visual question-answering datasets for surgical scenarios. The results show superior performance compared to previous works, indicating the potential of our model to tackle more complex surgery scenarios. 
\end{abstract}

\section{Introduction}

The rapid advancements in AI have increasingly focused on developing versatile assistants that can effectively understand and interact with the world through multiple sensory modalities, such as vision \cite{li2022elevater} and language \cite{brown2020language}. This multi-modal approach harnesses the unique strengths of each channel, enhancing the AI's ability to perform a wide range of real-world tasks more accurately and efficiently \cite{askell2021general, li2024multimodal}. Despite significant progress with large language models (LLMs) like GPT-3 \cite{liu2021makes}, GPT-4 \cite{achiam2023gpt}, and open-source alternatives such as LLaMA \cite{touvron2023llama} and Vicuna \cite{chiang2023vicuna}, these models typically handle language tasks in isolation, limiting their potential in applications that require a comprehensive understanding of multimodal data. Recent efforts have attempted to bridge this gap by integrating visual comprehension within a single model, aiming to create a unified representation that captures both visual and linguistic information. For example, models such as LLaVA \cite{liu2024visual} and Video-LLaMA (\cite{zhang2023video}) utilize shared visual encoders to process images and videos. 

In the surgical applications, the ability to understand and process both images and videos is of paramount importance \cite{saab2024capabilities, li2024llava}. Surgical procedures generate a wealth of visual data, including static images and dynamic videos. While general-domain vision-language models have been successful, they are less effective in surgical contexts because surgical visual-text pairs differ significantly from typical web content. This discrepancy can cause general-domain visual assistants to act like laypersons, either avoiding surgical questions or providing incorrect or completely fabricated responses. Despite significant advances in surgery visual question answering (VQA), prior methods often treat the problem as a classification task (e.g., choosing among specific answers from the training set) \cite{kirtac2022surgical, valderrama2022towards}. As a result, conversational generative AI for surgical applications is often restricted to specific tasks.

In this paper, we present Surgical-LLaVA, a first attempt to extent multimodal instruction-tuning to the surgical domain for multimodal conversational assistant. Inspired by recent work in instruction-tuning, Surgical LLaVA uses GPT-3.5 to generate diverse surgical multimodal instruction-following data using image/video-pairs, and fine-tune a surgical domain vision-langauge model. Specifically, our paper contributed follows as: 

\begin{itemize}
\item We propose Surgical-LLaVA, a multimodal model capable of engaging in meaningful conversations about surgical scenarios. It combines the language understanding capabilities of LLMs with a pretrained visual encoder tailored for spatiotemporal representations of surgical procedures.
\item We present datasets consisting of high-quality surgical visual instruction pairs, generated through a scalable and diverse annotation framework specifically designed for the surgical scenarios.
\item We achieved superior performance compared to existing instruction-following agents  in video reasoning for surgery scenario and visual question-answering. 
\end{itemize}
\section{Related Work}

\textbf{Large Language Models} The emergence of LLMs such as GPT , LLaMA and OPT \cite{zhang2022opt} has led to a paradigm shift in the field of natural language processing. These models excel in language generation and in-context learning, and demonstrate the ability to understand complex tasks. The high adaptability and generalisability of LLMs has led researchers to fine-tune these models for optimal performance.

One of the key strategies in such research is instruction tuning. This approach focuses on improving the model's alignment with user intent and optimising the quality of its output. For example, InstructGPT \cite{ouyang2022training} and ChatGPT  use this technique to improve their ability to interact with a variety of dialogues and answer complex questions. This effective approach has recently been applied to open source models such as Alpaca \cite{peng2023instruction} and Vicuna, resulting in performance improvements. 

\textbf{Leveraging LLMs for Multimodal Understanding}
The recent advancements in multimodal understanding have been primarily driven by the integration of image-based vision models with LLMs. Pioneering contributions, such as Flamingo \cite{alayrac2022flamingo} and BLIP-2 \cite{li2023blip}, have demonstrated the power of leveraging web-scale image-text data and cross-modal alignment techniques to exhibit impressive capabilities in conversational and few-shot learning settings. 
Equally, noteworthy is the emergence of Large Language and Vision Assistant (LLaVA) \cite{liu2024visual}, a model derived from the LLaMa architecture, which capitalizes on GPT-4's language proficiency to generate multimodal instruction-following data. Through instruction tuning on the derived data, LLaVA has showcased promising multimodal chat capabilities, hinting at the scalability potential of such an approach. Furthermore, the InstructBLIP \cite{dai2024instructblip} model has demonstrated strong image-based dialogue capabilities through vision-language instruction tuning and innovative instruction-aware visual feature extraction. Inspired by these success, several medical vision-language model have been studied \cite{shu2023visual, yunxiang2023chatdoctor, wu2023pmc}. LLaVA-Med \cite{liu2024visual} fine-tuned from biomedical data to instruction-following data and achieved superior performance on a variety of prompts. 

\begin{figure*}[ht]
    \centering
    \includegraphics[width=0.8\textwidth]{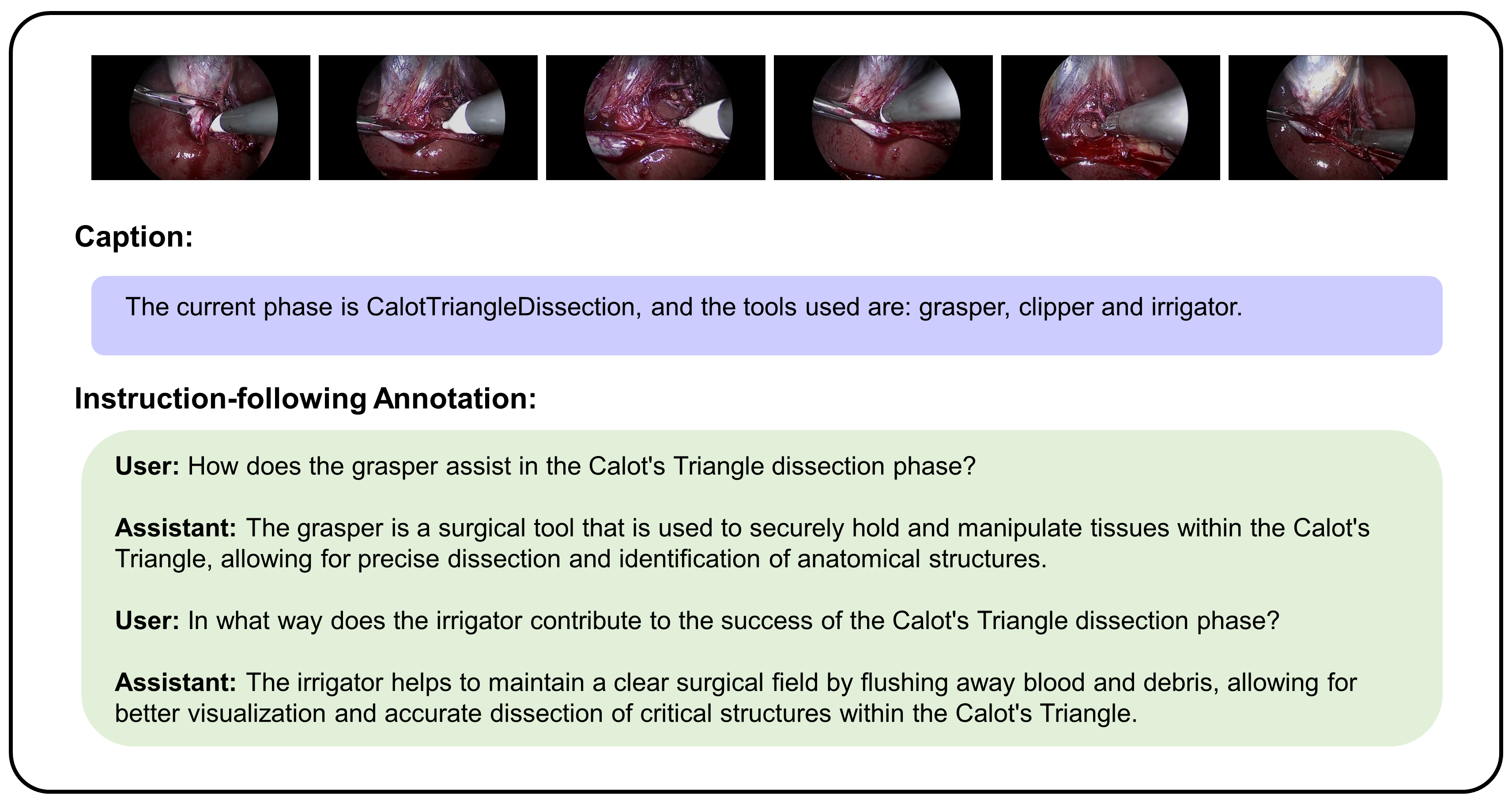}
    \caption{An example to illustrate the instruction-following data. We utilized the original caption to create an annotation that follows instructions with various prompts. The video and caption were acquired from Cholec80 dataset \cite{hong2020cholecseg8k}. The instruction-following data generated by GPT-3.5 using the text only (captions).}
    \label{fig:instruction}
\end{figure*}


\textbf{Surgical Scenario Visual Question Answering}
Early surgery video datasets primarily consisted of images and their corresponding annotations, focusing on tasks such as instrument detection, segmentation, and procedural step recognition. The Cholec80 dataset \cite{twinanda2016endonet} and the EndoVis18 dataset \cite{allan20202018} were pioneering efforts in this domain, providing annotated laparoscopic videos and surgical scenes for instrument recognition and segmentation, respectively. However, the creation and annotation processes for these datasets were labor-intensive and time-consuming, limiting their scalability and diversity.
To address these limitations, some researches shifted their focus towards leveraging the abundance of visual-text resources available in the medical domain. \cite{seenivasan2022surgical} and \cite{seenivasan2023surgicalgpt} pioneered the integration of visual and textual information by constructing datasets tailored for visual-question answering tasks in surgical settings. We aim to capture the rich multimodal information present during surgical procedures, enabling the development of models capable of simultaneously understanding and reasoning about complex visual and textual cues, thereby opening new avenues for research and allowing the exploration of novel tasks and applications that leverage the synergy between visual and textual information. Surgical-LLaVA aimed to develop an effective vision-language assistant for various complex prompts by generating multimodal instruction-following data for surgical scenarios by utilizing the language capabilities of LLMs such as GPT.

\section{Surgical Visual Instruction Data Generation}
This section describes a data-driven approach for multimodal instruction following data collection using LLMs within a novel framework specifically tailored to the surgical scenarios. Inspired by the recent success of visual language models in text annotation tasks, our approach is based on widely available image pair data, We adopted the LLaVA approach \cite{peng2023instruction} for data generation and incorporated annotation information as input to facilitate the generation of instructional data tailored to the surgical scenario. Specifically, our framework is the basis for generating a variety of contextualized instructions using expert-annotated surgical image data.\\
Recognizing the lack of comprehensive information in the original annotations, we attempted to leverage LLM's medical and background knowledge, such as GPT-3.5. We leveraged the original annotations to create instruction-following annotations with various prompts and instructions, as shown in Figure \ref{fig:instruction}.
By leveraging LLM's powerful language understanding and generation capabilities, it plays a key role in expanding the original annotations and incorporating relevant medical knowledge, procedural details, and contextual cues to create comprehensive and informative guideline-following annotations. To achieve this, we create a test set based on the ActivityNet-200 dataset \cite{caba2015activitynet} form, which contains videos accompanied by detailed descriptive captions and human-annotated question-answer pairs. Moreover, we construct an evaluation pipeline utilizing the GPT-3.5 model. The prompt used to generate the instruction data is provided in Appendix A.1. This approach not only allows us to generate high-quality, multimodal instruction data specific to surgical scenarios but also effectively utilizes existing annotation resources.

\begin{figure*}[h]
    \centering
    \includegraphics[width=0.75\textwidth]{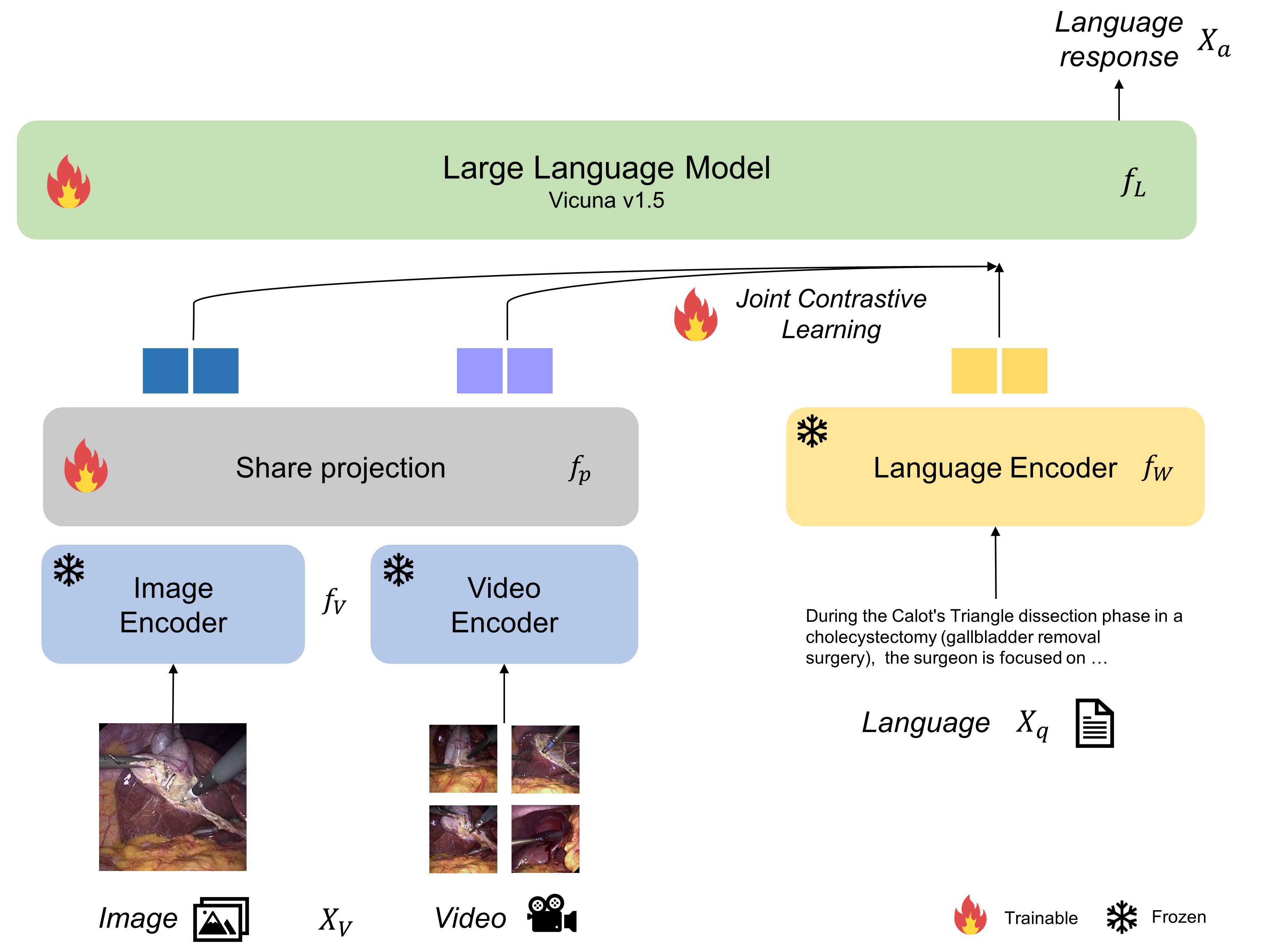}
    \caption{Architecture of Surgical-LLaVA. We adopted llava as the baseline, which vicuna as the LLM model and the pre-trained CLIP visual encoder ViT-L/14 as the visual model. The training involves encoding these inputs into token representations, followed by joint contrastive learning to align modalities within the semantic space.}
    \label{fig:pipeline}
\end{figure*}

\section{Surgical-LLaVA}
Surgical-LLaVA is a vision-language model that enhances surgical scenario analysis and conversation capabilities by aligning visual representations with a LLM. To achieve this, we leverage existing approaches used in the development of vision-language models for visual tasks. Given the scarcity of visual-caption pairs and the significant resources required for training from scratch, our strategy involves adapting pretrained image-based VL models for visual applications, as seen in previous works \cite{rasheed2023fine, ni2022expanding}. We specifically build upon the LLaVA, a large multimodal model that combines the visual encoder of CLIP \cite{radford2021learning} with the Vicuna language decoder \cite{chiang2023vicuna}, and is fine-tuned end-to-end on generated instructional vision-language data. We further fine-tune LLaVA with our visual-instruction data to tailor it for conversation tasks.

\subsection{Architecture}

The primary goal is to effectively apply the capabilities of the pre-trained LLM and visual model to surgical scenarios. The architecture is illustrated in Figure \ref{fig:pipeline}. We adopted LLaVA as the baseline, which vicuna as the LLM model and the pre-trained CLIP visual encoder ViT-L/14 as the visual model. Our visual encoder, originally designed for image processing, is extended to handle video inputs. 
Given a video sample \( V_i \in \mathbb{R}^{T \times H \times W \times C} \) with \( T \) frames, the encoder generates both temporal and spatial features. To derive video-level features, we perform average pooling on the frame-level embeddings along the temporal dimension, resulting in video-level temporal representations \( t_i \in \mathbb{R}^{N \times D} \). Similarly, average pooling along the spatial dimension produces video-level spatial representations \( z_i \in \mathbb{R}^{T \times D} \). By concatenating the temporal and spatial features, we obtain comprehensive video-level features.

\subsection{Visual Understanding Training}
The overall training process for Surgical-LLaVA follows a similar approach to LLM models like GPT. The model takes as input a text seqeunce $X_q$ and visual data $X_V$ (image or videos). These inputs are encoded into a token representation according to Eq \ref{eq1}. The training objective is to maximize the likelihood probability in Eq \ref{eq2}.
\begin{equation}
    \label{eq1}
    \mathbf{Z}_T = f_W(\mathbf{X}_T), \quad \mathbf{Z}_V = f_p(f_V(\mathbf{X}_V))
\end{equation}

\begin{equation}
    \label{eq2}
    p(\mathbf{X}_A \mid \mathbf{X}_V, \mathbf{X}_T) = \prod_{i=1}^{L} p_{\theta}\left( \mathbf{X}_A^{[i]} \mid \mathbf{Z}_V, \mathbf{Z}_T^{[1:i-1]} \right)
\end{equation}
where $L$ represents the length of the generated sequence, and $\theta$ denotes the trainable model parameters. This phase focuses on enabling the model to interpret visual representation from an extensive dataset comprising image/video-text pairs. Each visual sample corresponds to a single round of original caption data $(X_q, X_a)$, where $X_T = X_q$. \\

\textbf{Joint Contrastive Learning} In our approach, we employ a dynamic joint training that includes both image and video samples within each batch. 
 We employ a transformer model for our language encoder. The language encoder transforms these tokens into a text logit \( y \in \mathbb{R}^{L \times \mathcal{C}} \), where \( L \) is the length of the sequence. To align different modalities, we leverage contrastive learning techniques \cite{chen2020simple}. This approach aims to increase the similarity between paired data, bringing them into closer proximity within the semantic space, while decreasing the similarity between unpaired data. By using contrastive learning, we can associate each modality with the language component.

\begin{equation}
L_{\text{M2T}} = -\frac{1}{K} \sum_{i=1}^{K} \log \frac{\exp(x_i^\top y_i / \tau)}{\sum_{j=1}^{K} \exp(x_i^\top y_j / \tau)}
\end{equation}
In this context, \( x_i \) refers to the \( i \)-th modality data (image and video) and \( y_j \) to the \( j \)-th text, with both their features being normalized. The images and videos extracted from one video are treated as positive pairs, while images from different videos are treated as negative pairs. \( K \) stands for the batch size, and \( \tau \) is the temperature parameter. The temperature parameter  \( \tau \), set to 1 in our settings, controls the similarity distribution, balancing the focus between positive-negative pairs during training.

\subsection{Visual Instruction Tuning}
During instruction tuning, the model is trained to generate responses based on both visual inputs and text-based instructions. The input sequence for each conversation consists of text queries \( X_q \) and corresponding visual data \( X_V \), structured as follows:

\begin{equation}
    \mathbf{X}_T^r = \begin{cases} 
      X_q^1, &  r = 1 \\
      \text{Concat}(X_q^{r-1}, X_a^{r-1}, X_q^r), &  r > 1
   \end{cases}
\end{equation}
where $r$ represents the round number. For the first round, only the text query \( X_q^1 \) is considered. For subsequent rounds, the model concatenates the previous query-answer pairs with the current query to form the input. This allows the model to maintain conversation history across multiple rounds. The training objective is to maximize Eq \ref{eq2}, the same as in the previous step.


\section{Experiments}
\textbf{Implementation Details} We use LLaVA as our baseline model. We finetune the model for 3 epochs using a learning rate of 1e-5 and overall batch size of 16. The training of our 7B model took around 16 hours on 4 $\times$ RTX3090 24GB GPUs. During inference, for memory efficiency, we load the model in FP16 mode. 
The data in each batch is random combination of images and videos. \\

\textbf{Data Description} We utilized three datasets as visual datasets for our surgical scenario. 

\begin{itemize}
    \item Cholec80-VQA \cite{twinanda2016endonet} contains Q\&A pairs for 80 video sequences, including 97,251 Q\&A pairs on surgical phases and instrument presence of the Cholec80 dataset. The videos are configured at 25 frames per second (fps), while the annotations are provided at 1 fps. To align with the annotation frame rate, we extracted frames from the videos at 1 fps. We split the dataset into train/test sets (test set video sequences: 71-80).
    \item EndoVis-18-VQA \cite{allan20202018} 
    contains Q\&A pairs for 18 robotic Nephrectomy procedure video sequences, with 11,783 Q\&A pairs based on 2,007 surgical scenes from the MICCAI Endoscopic Vision Challenge 2018 dataset. For this dataset, we utilized 2,600 images and leveraged multiple annotations per single image. We followed the original train/test split of the EndoVis-18-VQA dataset.
    \item PSI-AVA-VQA \cite{valderrama2022towards} consists of 10291 Q\&A pairs with 35 answer classes (locations, surgical phases, and surgical steps) of holistic surgical scenario. They are constructed based on the surgical phase, step and location annotation provided in the PSI-AVA dataset. We followed the original train/test split of the PSI-AVA dataset.
\end{itemize}

\subsection{Surgical Video Understanding}
To evaluate the performance of Surgical-LLaVA on surgical scenario conversation, we present a benchmark designed to assess the text generation capabilities of visual models. The evaluation pipeline for video understanding follows Video-ChatGPT \cite{maaz2023video}. This pipeline evaluates the model's performance and assigns relative scores to the generated responses on a scale of 1-5, in the following three dimensions:
\begin{enumerate}[label=(\roman*), leftmargin=*, align=left]
    \item \textit{Conversation:} We assesses the accuracy and relevance of the model's responses during the visual dialogue, ensuring it accurately reflects the video content without any misinterpretations or false information.
    \item \textit{Detail description:} We evaluate the thoroughness of the model’s responses, checking for completeness by ensuring all major points from the video are covered, and for specificity by including precise details rather than generic statements.
    \item \textit{Complex reasoning:} We assess the model’s ability to engage in complex reasoning, ensuring its responses demonstrate an understanding of the video's context and logical connections between the content points.
\end{enumerate}

\begin{figure*}[h]
    \centering
    \includegraphics[width=0.75\textwidth]{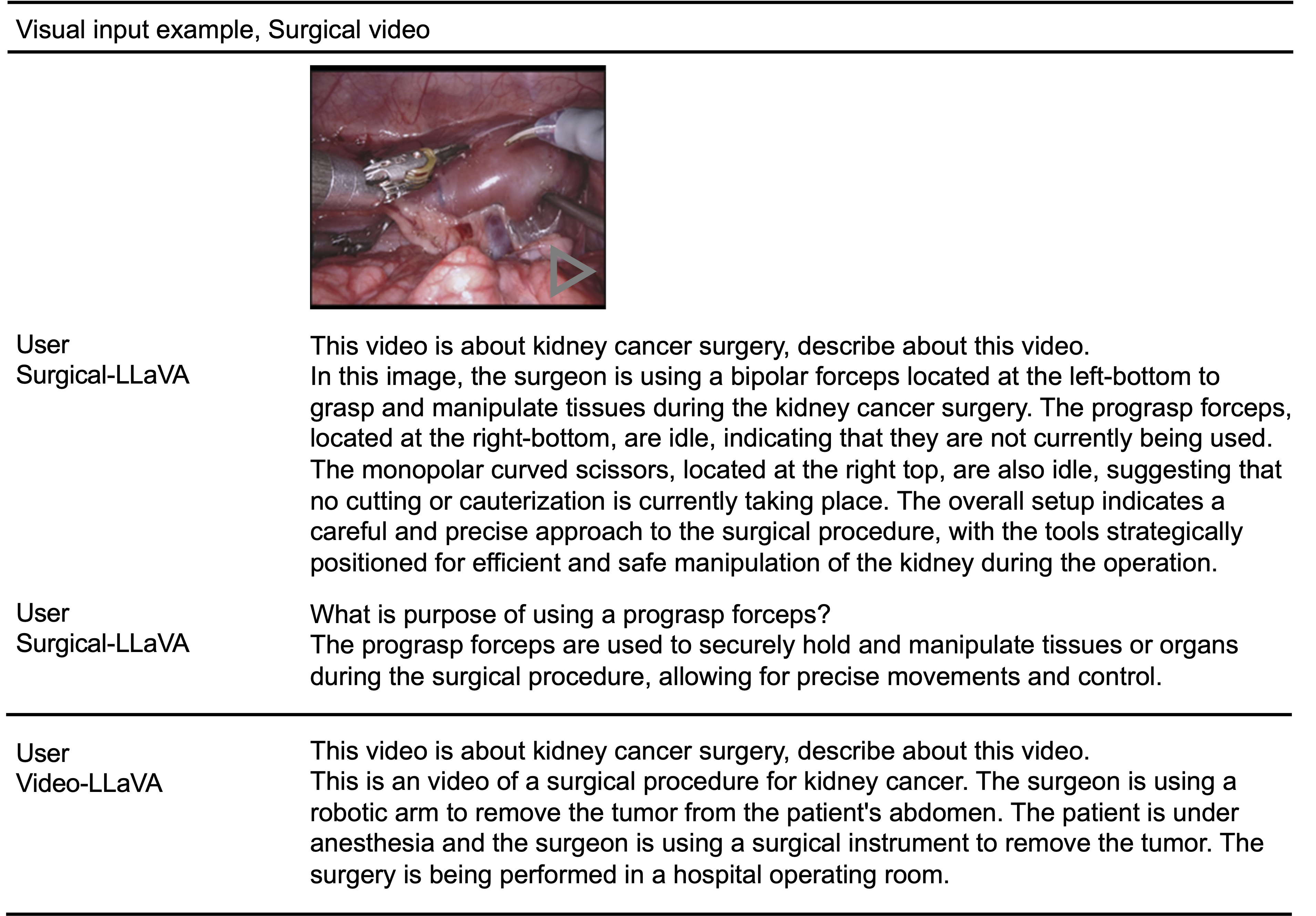}
    \caption{Example comparison of surgical visual chat and reasoning capabilities. Compared to Video-LLaVA \cite{lin2023video}, Surgical-LLaVA offers specific and accurate answers to surgical scenarios.}  
    \label{fig:visual sample}
\end{figure*}

Among the models evaluated, Surgical-LLaVA stands out with the highest scores across all three dimensions compared to other LVMLs that fine-tuned surgical instruction tuning data, as shown in Table \ref{table:ddd}. The Surgical-LLaVA model not only demonstrates superior conversational abilities and detailed descriptions but also excels in complex reasoning, particularly in understanding and articulating intricate surgical scenarios.  This capacity to grasp and reason through complex medical content is crucial, showcasing its potential for applications in surgical environments where accurate and nuanced interpretation of video content is paramount.
Figure \ref{fig:visual sample} illustrates examples of surgical visual conversations using different representative chatbots on images. Surgical-LLaVA responds to questions accurately, leveraging medical knowledge, whereas Video-LLaVA \cite{lin2023video} responds more like a layperson, often producing commonsense-based hallucinations. We used GPT-3.5 to compare Surgical-LLaVA to other LVLMs and to annotate the data. Additionally, when calculating the relative scores in Table \ref{tab:comparison1}, we consistently compared the answers of the candidate models to those of GPT-3.5. While the ranking of the resulting values is consistent, the values themselves may be biased in favor of GPT's answers. Given this self-enhancement bias, we expect Surgical-LLaVA's performance in practice to be more similar to GPT-3.5's than the current results suggest.
\begin{table*}[ht]
\centering
\caption{Comparison of large visual language models fine-tuned on surgical instruction data for video reasoning benchmarks.}
\label{table:ddd}
\resizebox{0.9\textwidth}{!}{%
\begin{tabular}{lcccccccc}
\toprule
\multirow{2}{*}{\textbf{Methods}} & \multirow{2}{*}{\textbf{LLM size}} & \multicolumn{2}{c}{\textbf{Conversation}} & \multicolumn{2}{c}{\textbf{Detail description}} & \multicolumn{2}{c}{\textbf{Complex reasoning}} \\
\cmidrule(lr){3-4} \cmidrule(lr){5-6} \cmidrule(lr){7-8}
& & \textbf{Accuracy} & \textbf{Score} & \textbf{Accuracy} & \textbf{Score} & \textbf{Accuracy} & \textbf{Score}\\
\midrule
Video-ChatGPT \cite{maaz2023video}& 7B & 42.7 & 3.1 & 38.0 & 2.6  & 39.8 & 2.5 \\
Video-LLaVA \cite{lin2023video}& 7B & 51.4 & 3.3 &  40.3 & 2.7  & 41.1 & 2.6\\
\rowcolor{cyan!10} Surgical-LLAVA & 7B & \textbf{58.3} & \textbf{3.9} & \textbf{47.1} & \textbf{3.2} & \textbf{46.5} & \textbf{3.1} \\
\bottomrule
\end{tabular}%
}
\end{table*}

\subsection{Evaluation on Visual Question-Answering Benchmarks}
In this evaluation, we assess the performance of various models on VQA tasks, particularly focusing on the Cholec80-VQA, EndoVis18-VQA, and PSI-AVA-VQA datasets. Table \ref{tab:comparison1} provides a comparative analysis of different models based on their performance metrics. 
The results in Table \ref{tab:comparison1} demonstrate that Surgical-LLaVA significantly outperforms existing models, achieving the highest accuracy rates across all three datasets. This ability to maintain superior performance across diverse datasets underscores Surgical-LLaVA's versatility and reliability in processing various types of visual and contextual information in surgical videos. The model's consistent excellence across multiple benchmarks represents a significant advancement in the field of surgical video data interpretation and interaction.
\begin{table*}[ht]
    \centering
    \caption{Comparison of various models on visual question-answering.}
    \resizebox{0.76\textwidth}{!}{
    \begin{tabular}{lccc}
        \toprule
        \textbf{Methods} & \textbf{Cholec80-VQA} & \textbf{EndoVis18-VQA} & \textbf{PSI-AVA-VQA}\\
        \midrule
        VisualBert \cite{li2019visualbert} & 89.7 & 61.4 & 58.5\\
        Block \cite{peng2020soloist} &89.5 & 60.1 & 59.9\\
        MFH \cite{yu2018beyond}&  87.5 & 58.8 & 47.8\\
        Surgical-VQA \cite{seenivasan2022surgical} & 89.8 & 63.2 & 65.6\\
        Surgical-LLaVA &  \textbf{92.2} & \textbf{68.7} & \textbf{67.1}\\
        \bottomrule
    \end{tabular}
    }
    \label{tab:comparison1}
\end{table*}

\subsection{Qualitative Evaluation}
\begin{figure*}[h]
    \centering
    \includegraphics[width=0.7\textwidth]{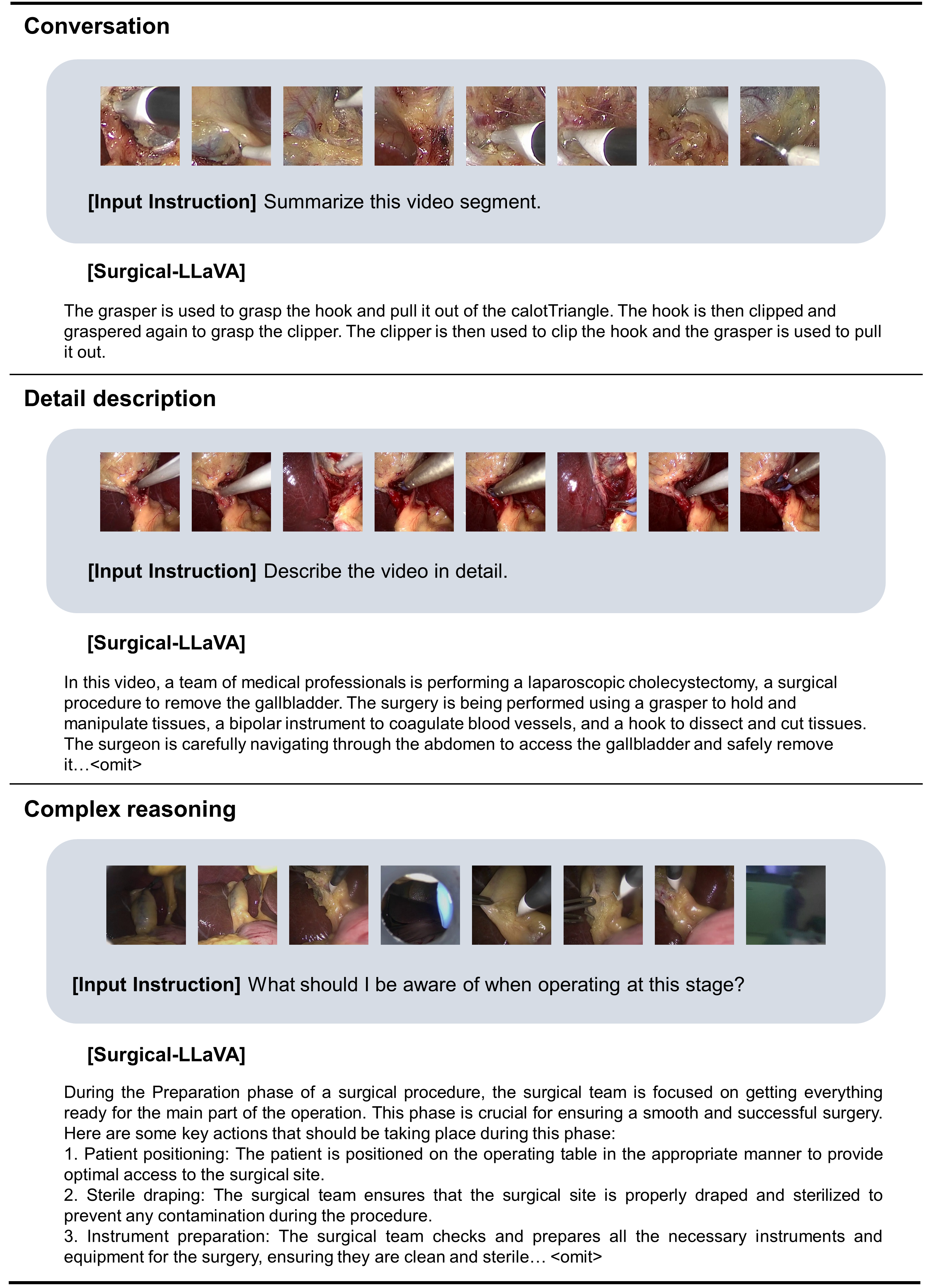}
    \caption{Examples from Surgical-LLaVA's demonstration of video reasoning. It shows conversation, detail description and complex reasoning cases.}
    \label{fig:detail}
\end{figure*}
To comprehensively assess the capabilities of our proposed Surgical-LLaVA model, we conducted an extensive qualitative evaluation spanning a diverse array of open-ended video question-answering tasks. \\
\textbf{Conversation}  We confirmed whether the model accurately reflects the content of the surgical videos without introducing any hallucinations or misinterpretations. This involves verifying that the generated text stays true to the visual information presented and is contextually appropriate as illustrated in top of Figure \ref{fig:detail}.\\ 
\textbf{Detail Description} We evaluated the model's capacity to generate detailed and descriptions of the surgical scenes. Surgical-LLaVAs describe the tools, steps, and even a description of the surrounding tissues in a surgery as illustrated in middle of Figure \ref{fig:detail}. \\
\textbf{Complex Reasoning} These tasks focused on the model's capability to perform complex reasoning based on the visual information and contextual knowledge. Surgical-LLaVA identified the current phase from the visual data and effectively suggest things to watch out for at that stage, as exemplified in bottom of Figure \ref{fig:detail}.\\
Throughout the evaluation, our Surgical-LLaVA model demonstrated remarkable proficiency in comprehending the visual content of the surgical videos and generating accurate, informative, and contextually relevant responses across the various tasks. The model effectively leveraged the visual information present in the videos to provide precise answers, detailed descriptions, and reasoned insights, showcasing its capability in understanding and reasoning about complex surgical procedures.

\begin{table}[hb]
\centering
\caption{Effect of joint training. We evaluate on three visual question-answering datasets. * denotes that we utilized only video data in both the first and second stages.}
\resizebox{0.75\textwidth}{!}{
\label{table:joint_training}
\begin{tabular}{lcccc}
\toprule
\textbf{Methods} & \textbf{Conversation} & \textbf{Detail description} &  \textbf{Complex reasoning}\\
\midrule
Surgical-LLaVA* & 57.5 & 44.5 & 42.0 \\
Joint with image & 58.3 & 47.1 & 46.5  \\
\(\Delta\) Acc. & +0.8 & +2.6  & +3.5 \\
\bottomrule
\end{tabular}}
\end{table}

\subsection{Ablation Study}
We conducted an ablation study on joint contrastive learning. As shown in Table \ref{table:joint_training}, we compared the performance of Surgical-LLaVA\textsuperscript{*} without image training. The model trained with both images and videos shows significant improvements across all metrics. These findings indicate that combining image and video training enhances the LLM's ability to comprehend visual representations in surgical scenarios.

\section{Limitations} 
The success of Surgical-LLaVA underscores the potential of combining large language models with specialized visual encoders for domain-specific applications. However, current public surgical datasets have limitations in providing limited information such as phase, tool and small amount of datasets. The ability to include specific and diverse information in surgical datasets will greatly improve scalability. 

The study should also examine how significant the instructional tuning data generated by the LLM is. We discussed qualitative results because we were unable to evaluate open-ended questions, and we need to quantitatively evaluate open-ended questions. To mitigate this, we are preparing to engage surgeons to create instructional captions for the videos and compare them with the responses from LVLM. This work is anticipated to provide valuable insights into multi-modal approaches for surgical scenarios within the LLM framework, paving the way for advancements in AI-assisted surgical training, decision-making processes, and patient care.

\section{Conclusion}
In this work, we introduced Surgical-LLaVA, a multimodal model designed for engaging in meaningful conversations and reasoning about surgical scenarios. By integrating the language understanding capabilities of LLMs with pretrained visual encoders tailored for spatiotemporal representations of surgical procedures, Surgical-LLaVA exhibits impressive multi-modal chat abilities in surgical contexts.
A contribution of our work is the introduction of a novel dataset consisting of high-quality surgical visual instruction pairs, generated through a scalable and diverse annotation framework specifically designed for the medical domain. Through quantitative and qualitative evaluations, we demonstrated Surgical-LLaVA's superior performance compared to existing state-of-the-art models in various tasks, including visual question-answering, video reasoning about surgical scenarios.

\bibliographystyle{plain}
\bibliography{ref}







\appendix

\section{Appendix}
\subsection{A.1. Instruction Tuning Data Generation Code}
We generated instruction tuning data for conversation, detailed description, and complex reasoning with the following prompt code \textit{"gpt info"} part with the original caption.

\begin{lstlisting}[language=Python, caption={Data Generation Code in Python}]
prompt = f"{gpt_info}\nBased on the following description, generate questions and answers that require complex reasoning to understand the scene."
prompt = f"{gpt_info}\nBased on the following description, generate a detailed description of the scene."
prompt = f"{gpt_info}\nBased on the following description, generate a question and answer that a human might ask about the scene."
\end{lstlisting}

\end{document}